\documentclass[a4paper]{svproc}
\usepackage{floatrow}
\newfloatcommand{capbtabbox}{table}[][\FBwidth]
\usepackage{blindtext}

\usepackage{cite}
\usepackage{graphics}
\usepackage{epsfig}
\usepackage[caption=false, font=footnotesize]{subfig}
\usepackage{multirow}
\usepackage{amsmath}
\usepackage{wrapfig}

\usepackage{url}

\newcommand{\etal}{\textit{et al. }}

\begin{document}
\mainmatter              
%
\title{Towards Assistive Robotic Pick and Place in Open World Environments}

\titlerunning{Towards Assistive Robotic Pick and Place}  
%
\author{Dian Wang\inst{1} \and Colin Kohler\inst{1} \and Andreas ten Pas\inst{1} \and Alexander Wilkinson\inst{2} \and \\Maozhi Liu\inst{1} \and Holly Yanco\inst{2} \and Robert Platt\inst{1}}
\authorrunning{Dian Wang et al.} 
%
%
\institute{Northeastern University, Boston MA 02115, USA\\
\email{wang.dian@husky.neu.edu},\\
\and
University of Massachusetts Lowell, Lowell MA 01854, USA}

\maketitle              

\begin{abstract}
Assistive robot manipulators must be able to autonomously pick and place a wide range of novel objects to be truly useful. However, current assistive robots lack this capability. Additionally, assistive systems need to have an interface that is easy to learn, to use, and to understand. This paper takes a step forward in this direction. We present a robot system comprised of a robotic arm and a mobility scooter that provides both pick-and-drop and pick-and-place functionality for open world environments without modeling the objects or environment. The system uses a laser pointer to directly select an object in the world, with feedback to the user via projecting an interface into the world. Our evaluation over several experimental scenarios shows a significant improvement in both runtime and grasp success rate relative to a baseline from the literature~\cite{openworldassistive}, and furthermore demonstrates accurate pick and place capabilities for tabletop scenarios.
\keywords{Assistive Robotics, Grasping, Human-Robot Interaction}
\end{abstract}

\section{INTRODUCTION}

\begin{figure}[t]
    \centering
    \subfloat[]
    {
    \includegraphics[height=2.5in]{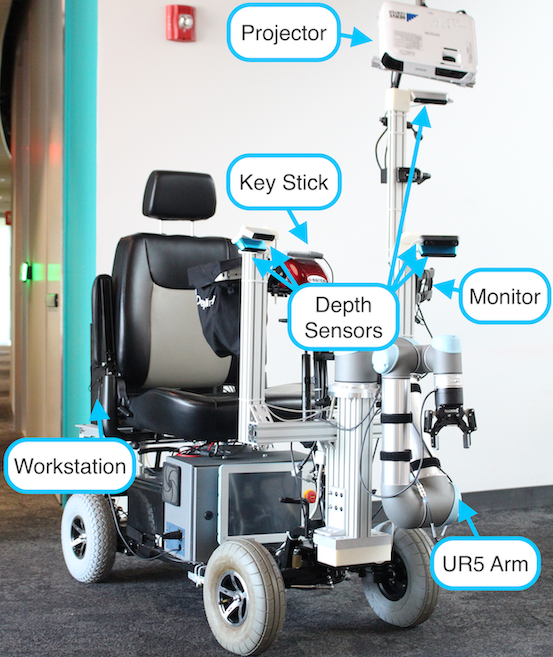}
    \label{system}
    }
    \subfloat[]
    {
    \includegraphics[height=2.5in]{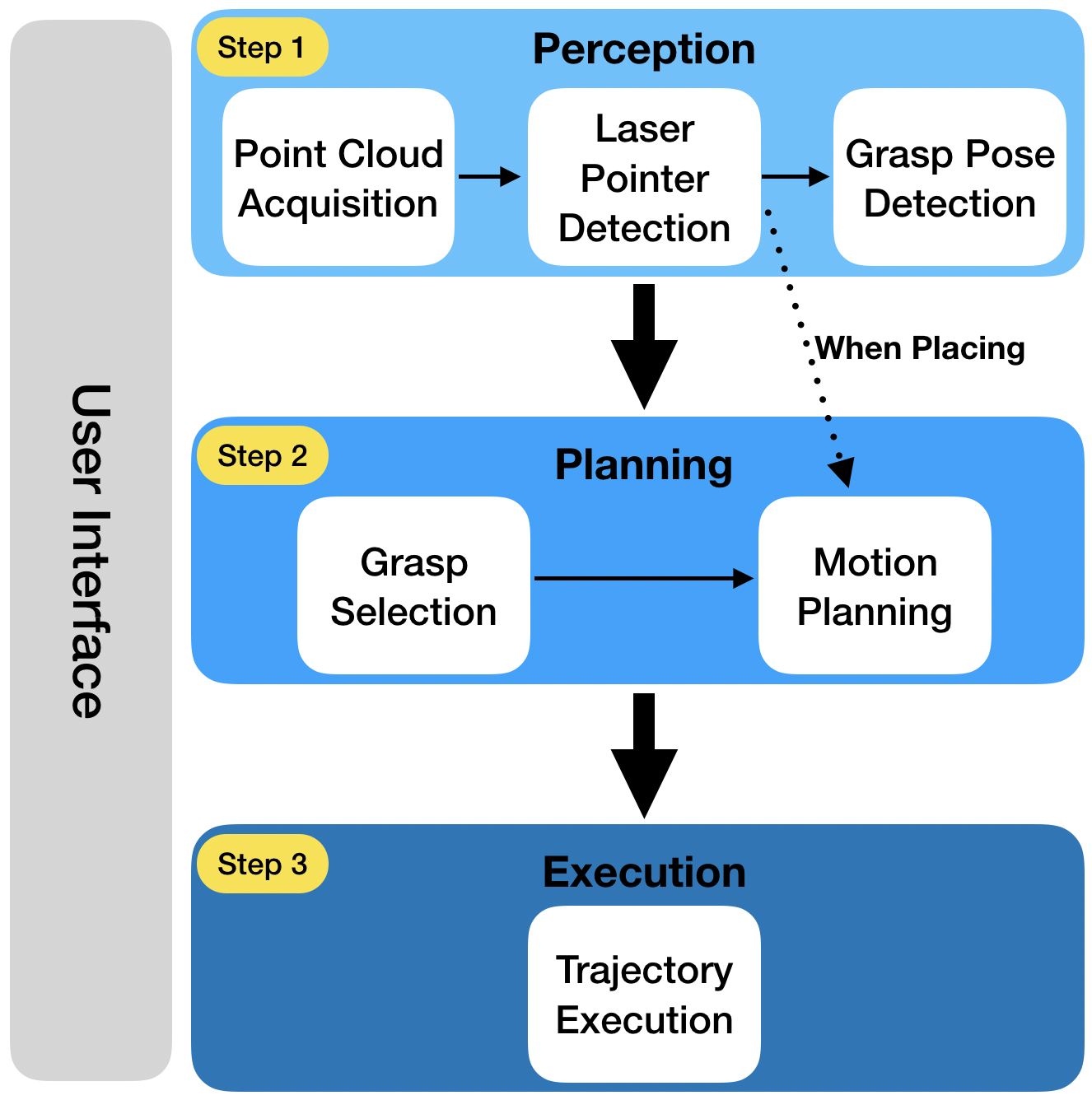}
    \label{fig:flow_chart}
    }
    \caption{
    (a) The system has a Universal Robotics UR5 robot arm mounted on a Merits Pioneer 10 mobility scooter. Five Occipital Structure depth sensors provide perception functionality for the system. The user interface includes a monitor, a key stick, a projector, and a dual laser pointer device. (b) The overview of our software workflow.}
\end{figure}

The ability to autonomously pick and place a wide range of novel objects can benefit applications in assistive robotics where robots are deployed to support humans in their daily life. The potential for this capability is enormous, given the continuously growing elderly populations in many countries~\cite{elderly} and the millions of people currently living with disabilities which affect motor functions~\cite{disabilities}. This paper proposes and evaluates a new robotic mobility scooter system (Fig.~\ref{system}) comprised of a standard mobility scooter augmented with a robotic arm that can perform novel-object pick-and-place tasks semi-autonomously. The system is operated by a human driving the scooter who identifies grasp objects and place targets using a laser pointer. Compared to a prior version of this system~\cite{openworldassistive}, the current system can perform pick-and-place tasks in domestic settings almost twice as fast, much more consistently, and with higher task success rates. Moreover, whereas the system reported in~\cite{openworldassistive} is only able to pick up objects, the new system can also place grasped objects using a similar laser interface as used during grasping. Finally, we provide a much more extensive evaluation of the system (under expert human control) than was provided in~\cite{openworldassistive}. 
The performance improvements described above and an improved user interface are due to a number of new system features including the following. 
1) A new strategy for manually selecting grasps found via grasp detection using a toggle interface. This strategy significantly reduces wrong-object grasping errors in cluttered environments. 2) A new type of user interface for assistive technology that utilizes dynamic spatial augmented reality (DSAR) to communicate grasp intent to the user by projecting light directly onto the real world. 3) A new configuration of five depth sensors that provides good coverage of the area in front of the vehicle without using metric SLAM. This makes sensing more reliable because it eliminates SLAM registration failures and it makes planned manipulator motions more reliable because the system observes a larger portion of the work space. 4) Use of stiff rather than compliant manipulator hardware. Much prior work in assistive manipulation uses compliant manipulators for safety~\cite{openworldassistive, assiitiveMobileManipulation}. 



\section{RELATED WORK}

The focus of research on assistive robotic systems for manipulation tasks has mainly been on human-robot interaction (HRI). Existing systems typically lack the ability to autonomously grasp and place an unknown object in an unstructured environment. Martens \etal introduced a semi-autonomous robotic system, composed of a robot arm mounted on a wheelchair with a speech interface~\cite{FRIEND}. Their approach used visual servoing techniques to grasp known objects. Kemp \etal developed a system with a laser pointer interface to allow a user to select an object in the world that their mobile robot would retrieve~\cite{pointandclick, EL-E}, with 88.9\% (32/36) grasp success rate in experiments on two different surfaces combined. However, the objects were placed in isolation, and only top-down grasps where allowed.
Achic \etal proposed the combination of a hybrid brain-computer interface and an electric wheelchair equipped with a robotic arm to facilitate navigation and manipulation tasks~\cite{hybridBCI}. Their experiments were restricted to pre-programmed robot arm motions. Pathirage \etal proposed a similar system but relied on a database of object models for grasping~\cite{Pathirage2013}. Grice \etal demonstrated an assistive robotic system, in which a PR2 robot could be teleoperated for manipulation tasks via a web browser and a single-button mouse~\cite{assiitiveMobileManipulation}. However, this required to control the end-effector directly which makes it much harder to use than autonomous grasping. Gualtieri \etal presented a system comprised of a Baxter robotic arm mounted on a mobility scooter which used a laser pointer device to locate the target object~\cite{openworldassistive}. 
Although the system was able to grasp novel objects, there were two primary issues: low grasp success rates and long task execution times.

Since novel object grasping is critical in domestic applications, grasp detection is a key element of our system. Several successful grasp detection systems have been recently proposed in the literature including the following. Mahler \etal proposed DexNet, another grasp system with a 93\% grasp success rate~\cite{dex-net2.0}. That system takes depth images as input and detects grasps in the plane, i.e. with a single degree of orientation freedom (orientation about the gravity axis). Morrison \etal presented a closed loop system that can grasp dynamically moving objects~\cite{closeloopgrasping}. Their algorithm runs at 50 Hz and reached grasp success rate of 83\% on unseen objects. Similarly, Viereck \etal proposed another closed loop grasping system for dynamically moving objects with a 88.9\% grasp success rate~\cite{pmlr-v78-viereck17a}. Kalashnikov \etal developed a reinforcement learning grasp system \cite{qtopt} using large amounts of real robot experience that generalizes to 96\% grasp success on unseen objects. Gualtieri \etal proposed GPD~\cite{graspposedetection, highprecision} that takes point cloud data as input and produces 6-DOF grasp poses as output. That system has a 93\% grasp success rate for novel household objects. In this paper, we decided to build on GPD because it was easier to adapt to unstructured domestic settings. In particular, since GPD operates with point cloud input, our system benefits from low-occlusion point cloud data provided by our five depth sensors. In addition, since GPD is not limited to detecting planar grasps, it can more easily generate side grasps -- an important capability in domestic settings.




\section{SYSTEM OVERVIEW}

\begin{figure}[tb]
    \centering
    
    \subfloat[]
    {
        \includegraphics[height=2.4in]{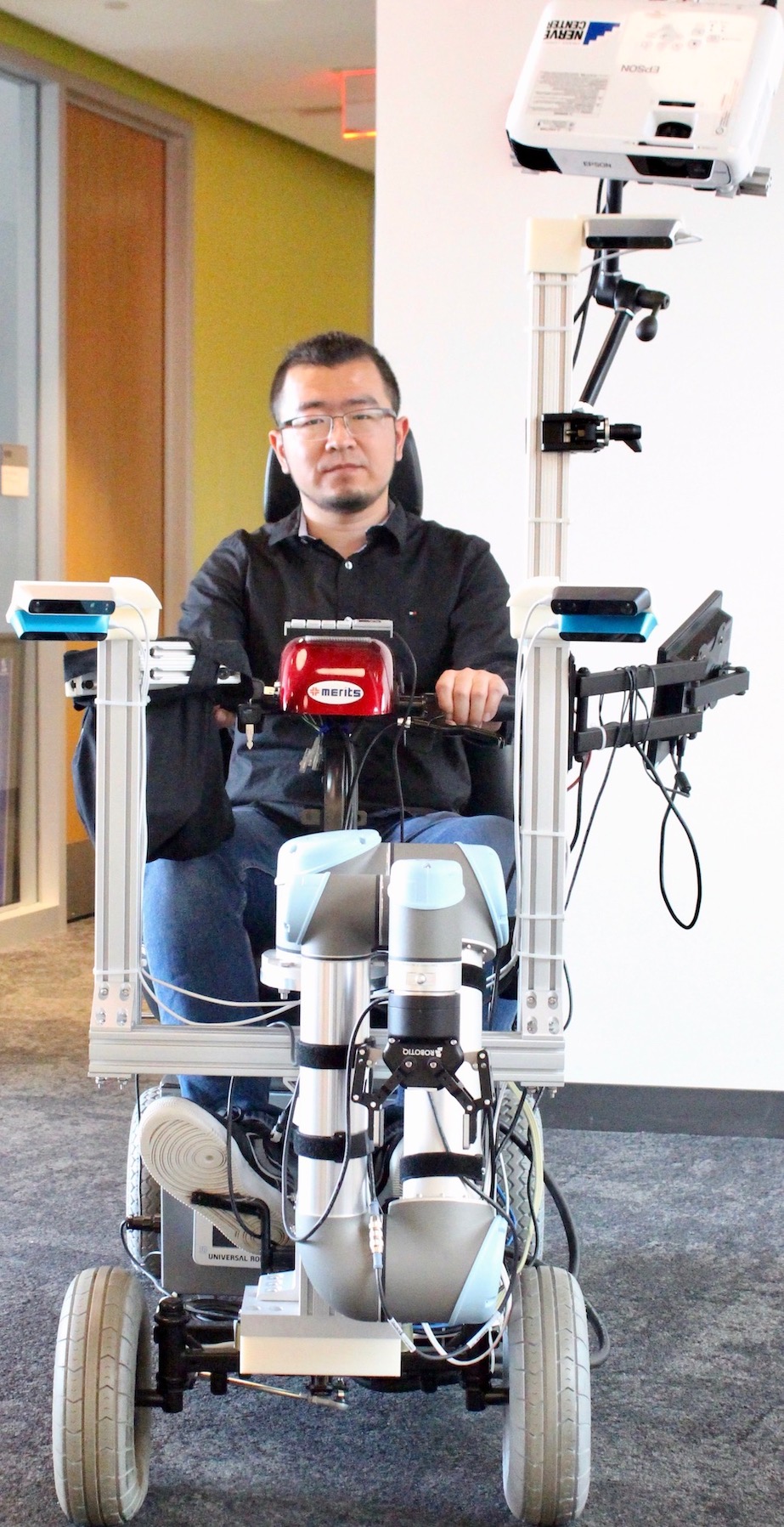}
        \label{fig:front}
    }
    \subfloat[]
    {
        \includegraphics[height=2.4in]{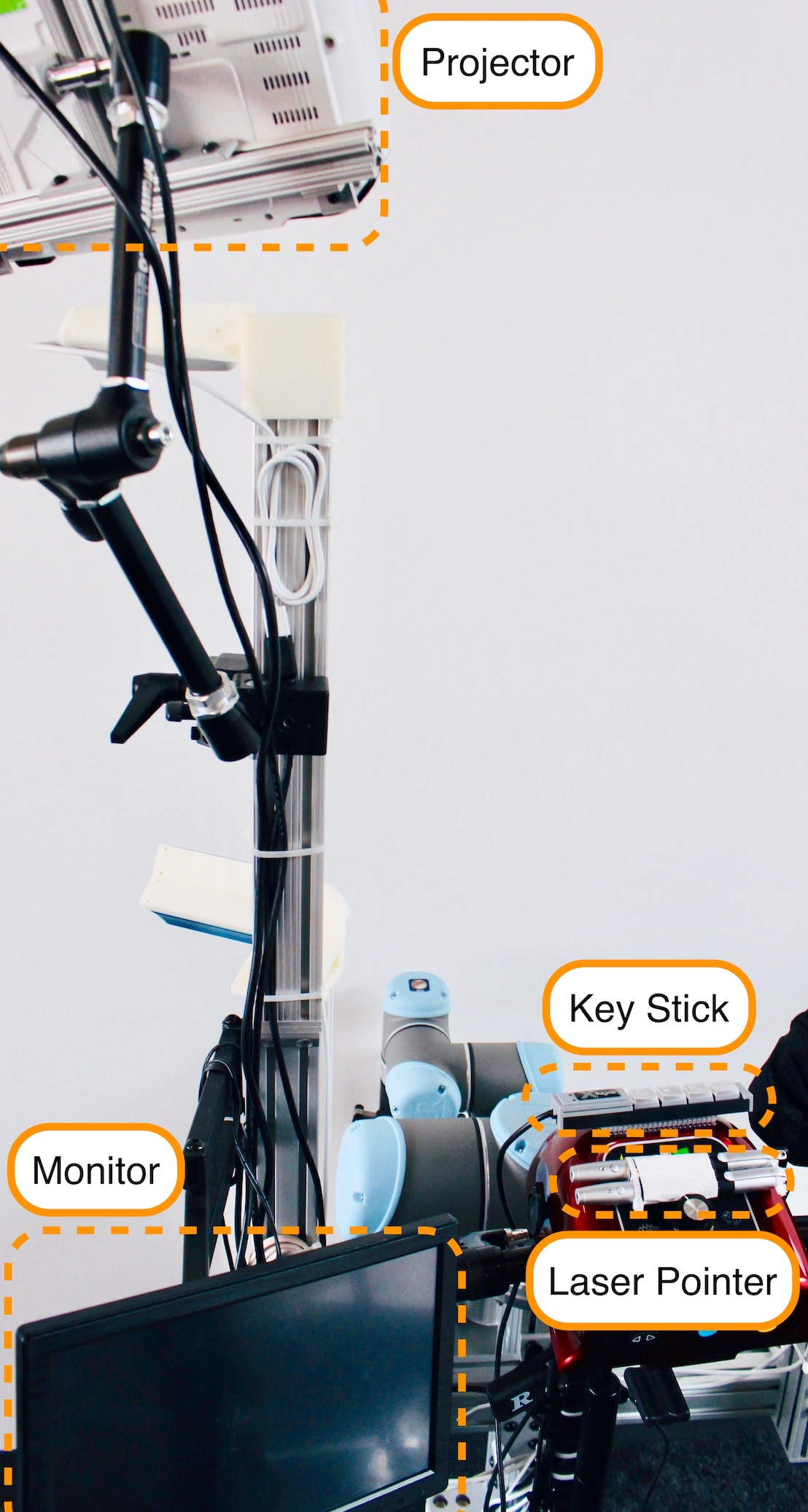}
        \label{interface}
    } 
    \subfloat[]
    {
        \includegraphics[height=2.4in]{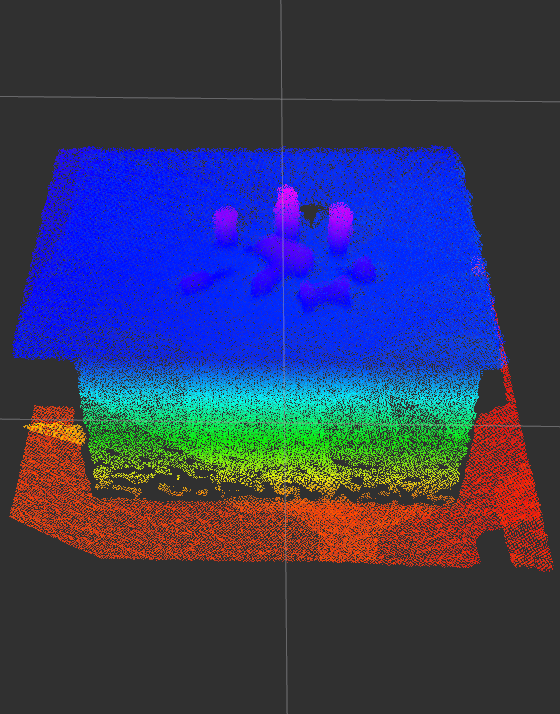}
        \label{fig:pointcloud}
    }
    \caption{(a) The front view of our system. (b) The user interface. (c) A nearly unoccluded point cloud generated by combining information from the five depth sensors.}
\end{figure}

Our system is comprised of a UR5 robotic arm (6-DOF), equipped with a Robotiq 2-finger 85 gripper, mounted on a Merits Pioneer 10 mobility scooter, as shown in Fig.~\ref{system}. The vision system is made up of five Occipital Structure depth sensors. Four are mounted near the handlebars (two on the left and two on the right) and an additional sensor is mounted higher up that can view elevated surfaces like the top of a shelf or a tall table. Using this five-sensor setup, the vision system can detect objects anywhere in a one-meter tall area in front of the scooter. As illustrated in Fig.~\ref{fig:front}, we mounted the sensors, their supporting structures, and the robot arm in such a way that the front view of the user is not blocked. This configuration also gives the sensors a nearly unoccluded view of the workspace: see the nearly complete point cloud in Fig.~\ref{fig:pointcloud}.

A dual laser pointer device, a ten-inch monitor, an X-keys XK-4 stick, and an EPSON LCD H801A projector make up the physical user interface (Fig.~\ref{interface}). A workstation is mounted on the back of the scooter and is connected to the robot, sensors, and user interface. The workstation consists of a 3.6 GHz Intel Core i7-6850K CPU (six physical cores), 32 GB of system memory, and a Nvidia GeForce GTX 1080 graphics card. 

\begin{figure}[t]
\newlength{\wflen}
\setlength{\wflen}{0.91in}
    \centering
    \subfloat[]
    {
        \includegraphics[height=\wflen]{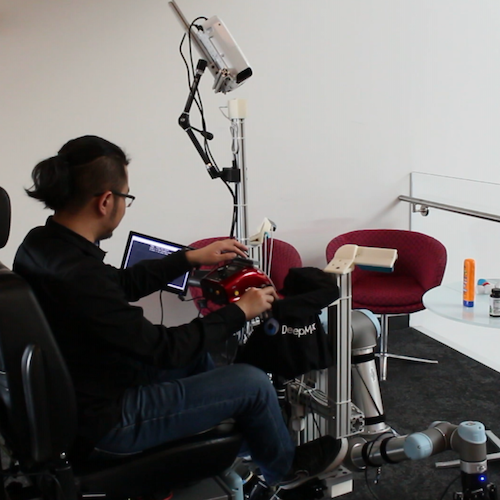}\hfil
    }
    \subfloat[]
    {
        \includegraphics[height=\wflen]{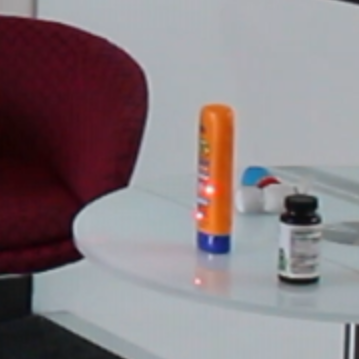}\hfil
    }
    \subfloat[]
    {
        \includegraphics[height=\wflen]{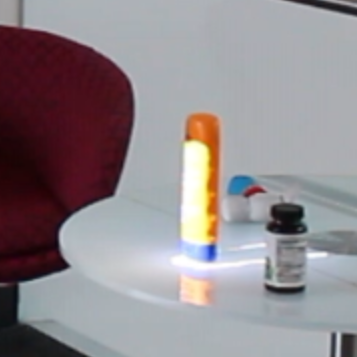}\hfil
        \label{fig:illuminating}
    }
    \subfloat[]
    {
        \includegraphics[height=\wflen]{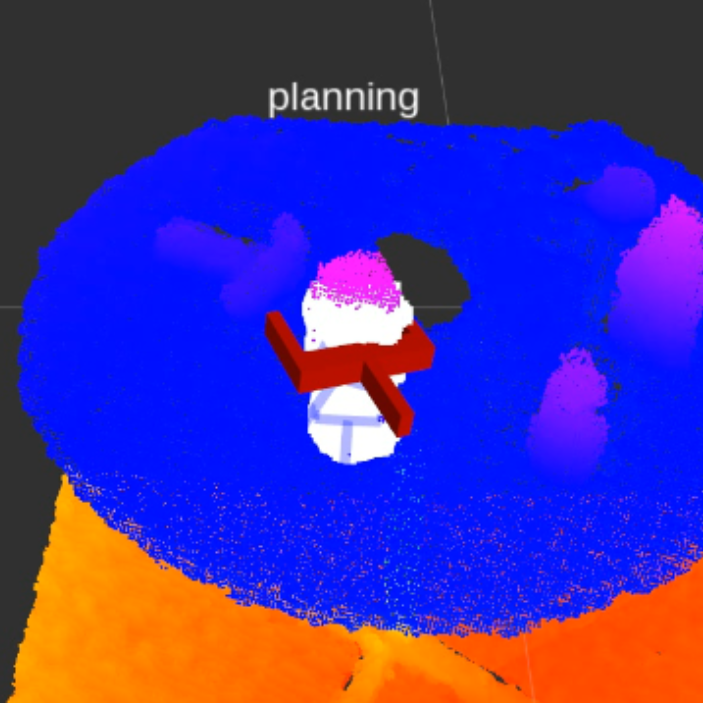}\hfil
    }
    \subfloat[]
    {
        \includegraphics[height=\wflen]{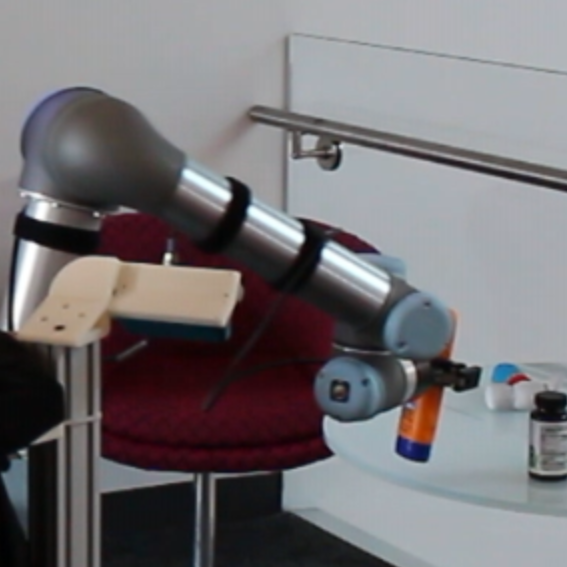}\hfil
    }
    \caption{The picking workflow of our system. (a) User drives to pick location. (b) User identifies target object using laser pointer. (c) Projector illuminates system grasp intent. (d) System detects grasp and solves motion plan. (e) Arm executes grasp.}
    \label{fig:picking}
\end{figure}

The software of our system can be divided into subsystems for perception, planning, execution, and user interface, illustrated in Fig.~\ref{fig:flow_chart}. 
The perception subsystem consists of three parts. First, we acquire a fused point cloud by concatenating the individual point clouds from all five depth sensors. Second, we detect the 3D position of the laser pointer using the infrared radiation (IR) images. Third, we detect potential grasps using the Grasp Pose Detection (GPD) library~\cite{highprecision, graspposedetection}. Once grasps have been detected, the planning subsystem selects a grasp based on either automatic grasp selection or manual grasp selection provided by our system, and then tries to find a feasible trajectory for the robot arm. The execution subsystem executes that trajectory. When performing placing, grasp pose detection and grasp selection are skipped and the planning subsystem will directly generate a trajectory to the position that the laser pointer indicated. The user interface subsystem interacts with the user during the whole process. All subsystems are implemented as ROS~\cite{ROS} nodes running on the workstation. We describe the software subsystems in the following sections. Fig. \ref{fig:picking} shows the process of our system picking up an object.

\section{USER INTERFACE}
We developed a simple graphical user interface (GUI) using RViz~\cite{ROS}. The GUI is used to visualize the various outputs of our system: the perception results, the status of the system, and the current actions available to the user. The user interacts with this interface by using the key stick. The dual laser pointer allows the user to select the next target (either the object to be grasped or the location for placing) for the system. The laser device consists of two (5 mW 650 nm) red lasers. The user positions the pointer by illuminating only a single laser. When the laser is positioned as desired, the user activates the second laser to confirm the selection.

 
A dynamic spatial augmented reality (DSAR) system augments our user interface by providing more informative feedback using a projector mounted on the scooter (Fig. \ref{system} and \ref{interface}).
This system highlights an area with a beam of light indicating the target of the next pick or place action.
Specifically, while picking the area around the selected object will be highlighted (Fig \ref{fig:illuminating}) and  
while placing the position selected for placing an object will be highlighted. 

To do this, we first create a virtual camera that has the same intrinsics and extrinsics as the projector. 
We then project the segmented point cloud into the frame of the virtual camera to create a black-and-white image where pixels are white if the corresponding point belongs to the segmentation and black otherwise. Finally, the image is sent to the projector and the white pixels are illuminated.


In order to determine the intrinsics of the projector lens, the projector was mounted a fixed distance away from a wall and an image was projected upon it.
The projector was then modeled as a pinhole camera with an intrinsic camera matrix $K$:
\begin{equation}
K=
\begin{bmatrix}
	f_x & 0   & c_x \\
	0   & f_y & c_y \\
	0   & 0   & 1
\end{bmatrix}
\end{equation}
Let $W$ and $H$ be the width and height of the image respectively, $w$ and $h$ be the width and height of the image in pixels respectively, $X$ and $Y$ be the distances from the principal point (the intersection of the optical axis and the image) to the origin (top-left) point of the image on the image plane in their respective axes, and $Z$ be the distance from the projector to the image plane. We can determine the focal length $f$ of the projector in pixels using the projective equation: $fx = fy = w \frac{Z}{W}$.
The principal point $(c_x, c_y)$ is then calculated in pixels: $c_x = w \frac{X}{W}$; $c_y = h \frac{Y}{H}$.



\section{PERCEPTION}
\subsection{Point Cloud Generation}

Both the grasp detection and motion planning use point cloud information as the primary input, so it is important to obtain the most complete point cloud possible of the workspace.
Our system uses five Occipital Structure depth sensors. Four of them are mounted 114cm above the ground, and the other one is 156cm. The higher sensor points 44 degrees downward. The four lower sensors are separated into two mirrored groups. One in each group points 25 degrees downward which covers most of the workspace. The other one points 55 degrees downward which supplement the view of the area lower and closer to the arm. The distance between the two groups is 63cm. All five sensors also have an inward degree of 20. Coverage of the workspace with multiple views is important for both obstacle avoidance and for grasp detection; as shown by~\cite{highprecision} and~\cite{graspposedetection}, grasp detection accuracy can improve by as much as 9\% simply by adding one additional view. Our system attempts to provide at least two views of the workspace using the five-sensor setup: the sensors on the left and right group provide two views, and the higher one provides another view. Fig.~\ref{fig:pointcloud} shows an example of a point cloud generated from our cameras. The point cloud from our five cameras covers all of the reachable workspace of the robot arm, as is indicated by the complete table being seen in front of the robot.



\subsection{Laser Pointer Detection}

A dual laser pointer device is used to identify the target for the next pick or place action. We detect the position of the laser pointers by reading the infrared radiation (IR) image from the depth sensors. Since our device consists of two parallel laser pointers, each potential location is detected by finding two nearby high-value areas within each IR image using hierarchical clustering~\cite{clustering}. To increase the accuracy, we detect potential locations in the IR image of each of the five sensors and map them to 3D to find a matching position.

\section{GRASP DETECTION}

\subsection{Preprocessing}

The goal of point cloud preprocessing is to eliminate from consideration the parts of the environment that can be safely excluded using standard methods.
First, we remove all planes that are roughly horizontal with respect to the vehicle from the cloud. We use RANSAC~\cite{RANSAC} parameterized with a distance threshold of 0.015m and a maximum angle relative to the horizontal of five degrees. Planes are not removed unless they contain at least 13k inliers. We eliminate at most 70\% of the points in the cloud this way. After removing horizontal planes,
we attempt to segment individual objects using Euclidean cluster extraction~\cite{Semantic3DObjectMaps} with a distance threshold of 0.005m. We eliminate all clusters except the one closest to the point of interest found using our laser point detection subsystem. If the remaining cluster has fewer than 500 points, we skip this step and simply return all points within a 0.1m radius ball centered at the laser pointer position.

\subsection{Grasp Pose Detection}

After point cloud preprocessing is complete, we detect feasible grasps using 
GPD~\cite{graspposedetection,highprecision}.
GPD detects 6-DOF robotic hand poses that are predicted to be feasible grasps. The system takes two different point clouds as input: the segmented cloud as described above and the full cloud. GPD uses the segmented cloud to seed candidates for grasp detection and the full cloud to ensure that the detected grasps do not collide with the environment. Because the seeded grasps are drawn only from the segmented cloud, the system only detects grasps in the vicinity of the segmented object.

\section{GRASP AND PLACE SELECTION}

After grasp candidates are found, the system filters out all grasps that do not have a collision-free inverse kinematics (IK) solution~\cite{IK}. To select a grasp to execute from the set of remaining feasible grasps, the user can either defer to a built-in set of heuristics or select the grasp manually.

\subsection{Automatic grasp selection}
\label{sec:automatic}

In automatic grasp selection, we prioritize the grasps using an objective function equal to the product of the following five factors:
\begin{equation}
C = C_w C_j C_a C_s C_p
\end{equation}

$C_w$ penalizes grasps that are near the gripper width limits. Small errors in perception or kinematics can make grasps that are close to the maximum width of the gripper fail. Grasps that are close to the minimum width often correspond to small parts of an object that are not suitable for a mechanically stable grasp. Let $G_w$ be the grasp width, i.e., the extent of the points in the closing direction of the gripper, $W_{\rm min}$ be the min gripper width, $W_{\rm max}$ be the max gripper width, and $W_{d}$ be a minimum acceptable distance from the gripper extrema (0.015m in our experiments). $C_w$ is:
\begin{equation}
    C_w=1-\frac{\max(0, W_{d}-\min(|G_w-W_{\rm min}|, |G_w-W_{\rm max}|))}{W_{d}}
\end{equation}

$C_j$ gives preference to grasps that can be reached by moving the arm in short distances in configuration space. Grasps with larger joint distance need more execution time. Let $J_i$ be the initial arm configuration, $G_j$ be the arm configuration which reaches the grasp pose, and $J_m$ be the max joint distance. The joint distance cost $C_j$ is computed as:
\begin{equation}
    C_j = \max(0, 1 - \frac{1}{J_m} \| J_i - G_j \|)
\end{equation}

$C_a$ takes the approach vector of the grasp into account giving preference to top grasps over side grasps. When scoring side grasps, it prefers grasps that approach the object from the front. Top grasps are easier for the robot arm to execute. Side grasps that approach an object from the front are less likely to collide with objects blocked by the target which might not be observable by the sensors. Let $G_a$ be the approach vector of a grasp and $G_x$ be the axis vector. $C_a$ is computed as:
\begin{equation}
C_a=
\begin{cases}
0.5 + 0.5 |G_a[1]|, |G_x[2]| > 0.8\\
1, \text{otherwise}
\end{cases}
\end{equation}
where $G_a[1]$ is the $y$ component in the approach vector (the $y$ axis in the world frame is pointing forward) and $G_x[2]$ is the $z$ component in the axis vector.

$C_s$ encodes a preference for grasps that are nearby the segmented cloud as this decreases the possibility that the object will slide away during grasping. Let $G_b$ be a fixed point relative to the grasp position, $l$ be the gripper length, and $S$ be the sample cloud. The sample distance cost $C_s$ is computed as:
\begin{equation}
    C_s = 1-\frac{\min(l, \min_{\forall p \in S}\|p - G_b\|)}{l}
\end{equation}

Finally, $C_p$ penalizes grasps far away from the detected laser pointer position. Let $p$ be the laser pointer position and $\rm{th}$ be a distance threshold (0.05m in our experiments). The position distance cost $C_p$ is computed as:
\begin{equation}
    C_p=\exp(-10*\max(0, \|p-G_b\|-\rm{th}))
\end{equation}

\subsection{Manual Grasp Selection}
\label{sec:manual}

Although automatic grasp selection often works well, there are a few scenarios where it has problems. First, in densely cluttered scenes, the segmented point cloud might contain multiple objects and a grasp on the wrong object is selected. Second, the system may attempt to grasp the table, shelf, or some other part of the environment. Third, in the case of pick-and-place, the user might have a preference for grasping the object in a particular way. The option for manual grasp selection enables the user to overcome these problems by specifying exactly how to grasp the object. Note that the grasp detection system is still active here -- the only difference is that after grasp detection has occurred, the user may select which detected grasp is most suitable instead of relying on the heuristics described earlier.

During manual grasp selection, the set of feasible grasps are visualized on a small monitor mounted near the scooter handlebars (see Fig.~\ref{system}) and the user can toggle through them using the key stick to select the desired one. To facilitate this process, we cluster the detected grasps to remove nearby duplicates using hierarchical clustering~\cite{clustering} on the 3-DOF position of the grasps. The maximum Euclidean distance inside each cluster of grasps is 0.02m. We then display the clusters. 

\subsection{Place Selection}

When in pick-and-place mode, grasps are generated and executed in the same manner as above. However, after the grasp is selected, we measure the distance between the grasp and the surface the object is on. This offset will be added to the next laser pointer detection in order to adjust to the height the object was grasped at. Once the grasp has been executed, the user is free to re-position the scooter in front of the surface they wish to place the object onto. The placement pose is generated by combining the orientation the object was grasped in and the position indicated by the second laser pointer detection.

\subsection{Motion Planning}

OpenRAVE~\cite{openrave} is used for collision checking and motion plan generation. There are two steps when computing motion plan. First, linear trajectories are attempted. If no collision-free trajectory can be found, trajectory optimization, TrajOpt~\cite{trajopt}, is used. When grasping, the system iterates through the grasps in sorted order (based on Section~\ref{sec:automatic} and~\ref{sec:manual}) and tries to generate a collision-free motion plan for the whole grasping task, i.e., reaching, grasping, lifting, and dropping. We execute the first grasp for which a valid motion plan can be found.

\section{EXPERIMENTS}


\begin{figure}[t]
    \centering
    \subfloat[]
    {
        \includegraphics[height=1.65in]{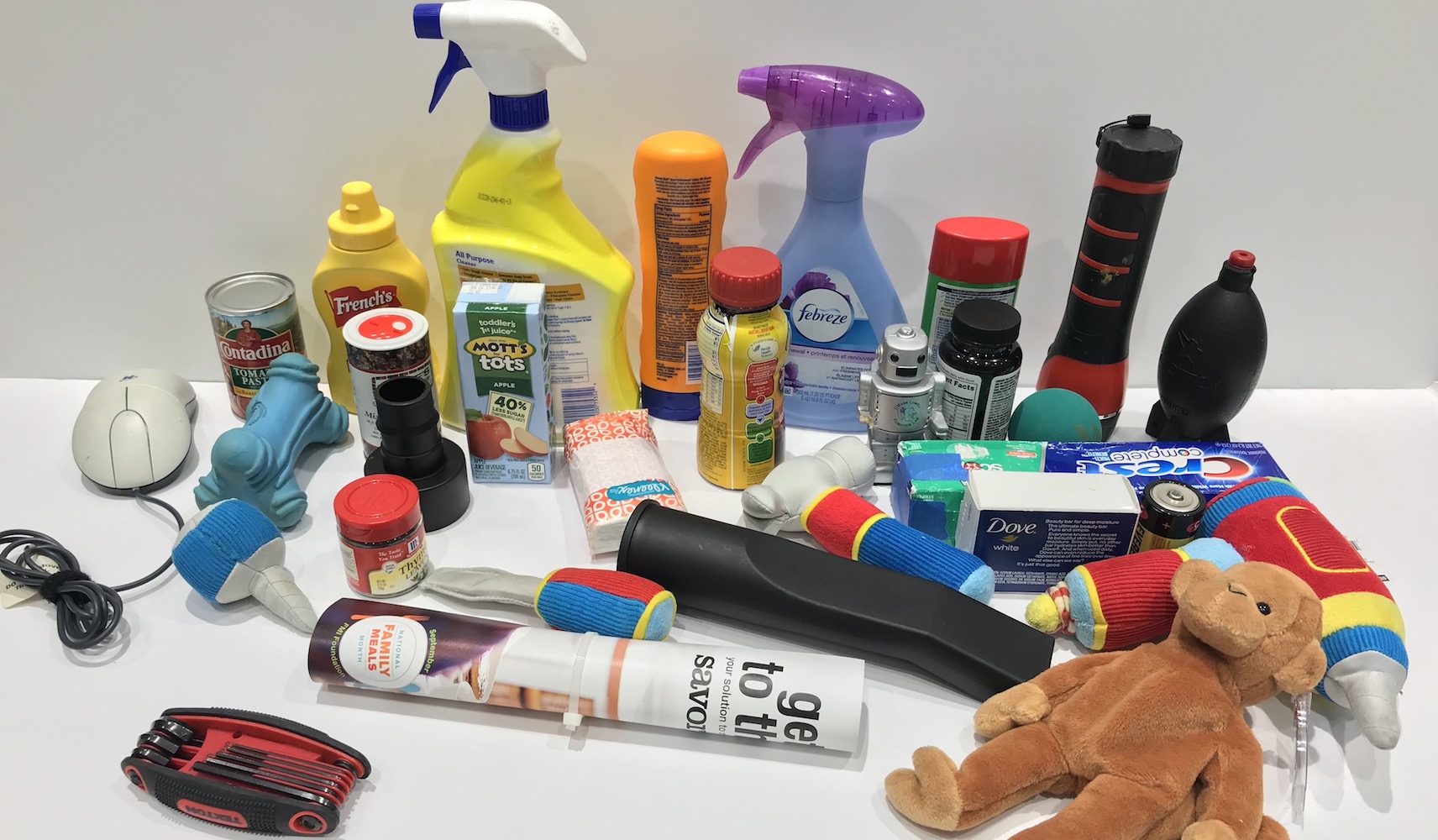}
        \label{objects}
    }
    \subfloat[]
    {
        \includegraphics[height=1.65in]{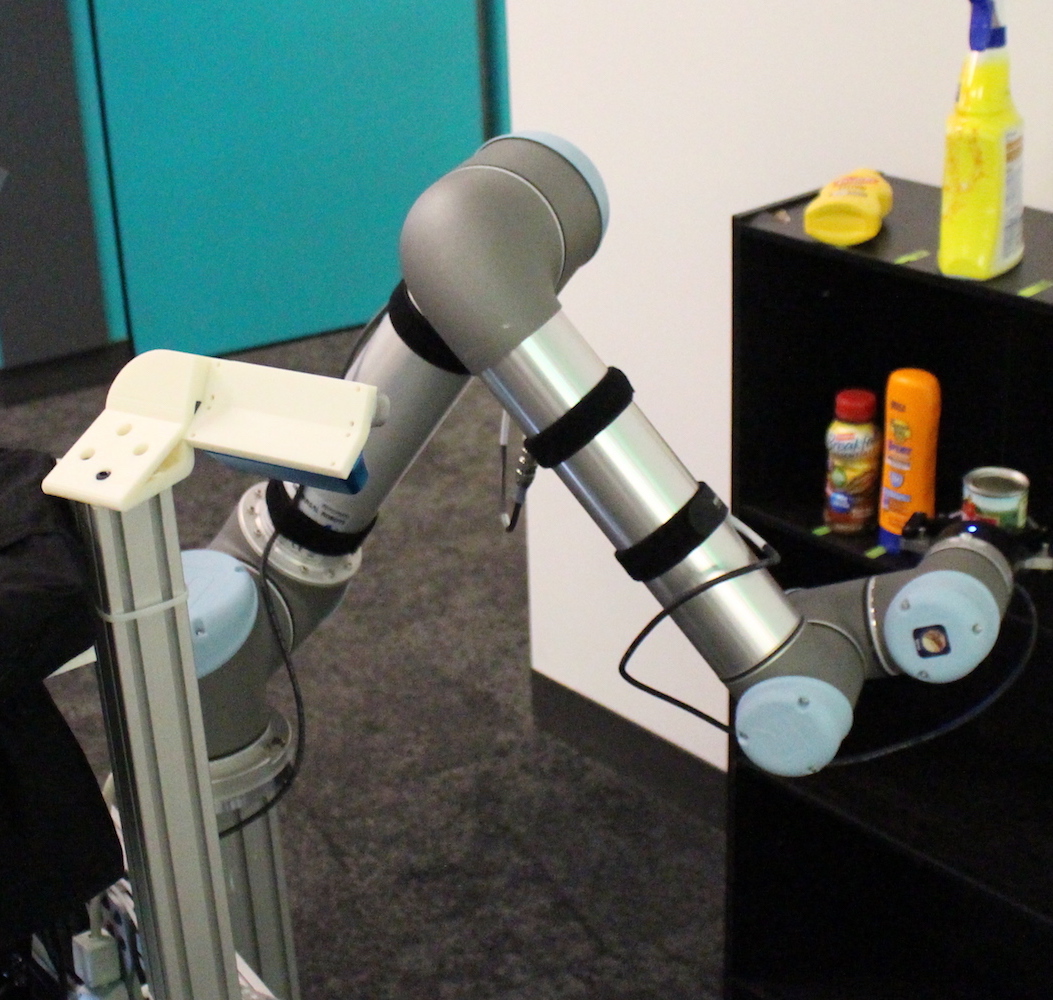}
        \label{fig:exp_open}
    }
    \caption{(a) The 30 objects used in the experiments. (b) Our system grasping in-situ.}
\end{figure}

We performed experiments that compare the performance of our system to the results from~\cite{openworldassistive}. Then we performed an additional experiment that evaluates pick-and-place functionality in a more challenging environment. 
We also compare our automatic grasp selection against manual grasp selection.
Fig.~\ref{objects} shows the 30 objects used to test the system. These objects are similar to those used in Gualtieri \etal~\cite{openworldassistive} to allow a fair comparison. 

\subsection{Evaluating Grasping in Isolation}
\label{sec:exp1}
\begin{figure}[b!]
\begin{floatrow}
\capbtabbox{%
\scriptsize{
    \begin{tabular}{|c|c|c|}
    \hline
    & {Work of~\cite{openworldassistive}} & {This Work} \\
    \hline
    {Grasp SR} & 89.6\% (78/87) & 93.8\% (90/96) \\
    \hline
    {Detect SR} & 87.8\% (108/123) & 92.3\% (96/104) \\
    \hline
    \end{tabular}
    }
}{%
    \caption{Results for Grasping In Isolation}
    \label{tab:exp1_result}
}
\capbtabbox{%
    \scriptsize{
    \begin{tabular}{|c|c|c|}
    \hline
    & {Avg.} & {Std.}\\
    \hline
    {Acquire Point Cloud} & 8.0s & 2.4s\\
    \hline
    {Detect Laser Pointer} & 5.8s & 5.3s \\
    \hline
    {Detect Grasp Pose} & 7.4s & 2.3s\\
    \hline
    {Grasp Filter} & 1.6s & 0.6s\\
    \hline
    {Plan Motion} & 2.4s & 2.8s\\
    \hline
    {Execute} & 24.2s & 1.1s\\
    \hline
    {Attempt Total} & 51.7s & 7.2s\\
    \hline
    \end{tabular}
    }
}{%
    \caption{Timing for Grasping In Isolation}
    \label{tab:exp1_time}
}
\end{floatrow}
\end{figure}
\subsubsection{Experimental Protocol}

To compare the grasping functionality of our system to prior work~\cite{openworldassistive}, we performed 15 trials in a tabletop scenario on a 46 cm tall table while the scooter was stationary. For each trial, six objects were randomly selected from the object set and placed on the table in randomly selected positions at least two centimeters away from each other in an upright orientation. These trials were run with automatic grasp selection in pick-and-drop mode by an expert user who determined the order the objects were grasped in. We evaluated the detection success rate, grasp success rate, and the detailed time spent. Detection success is defined as the selected grasp being on the desired object. Grasp success means that the robot was able to grasp and transport the object to a handlebar-mounted basket. When measuring the time the system took, we only average over successful detections, grasps, plans, etc. We do not include system loss time, e.g., time spent by the user to press a key.

\subsubsection{Results}

Table~\ref{tab:exp1_result} shows the results from this experiment. We attempted to remove all objects from the table 15 times. Each time, the table was initialized with six objects placed in random locations as described above (a total of 90 objects to be grasped). It took our system 104 attempts to grasp these 90 objects where 96 out of 104 (92.3\%) detection attempts succeeded and 90 out of 96 (93.8\%) grasping attempts succeeded. Compared to~\cite{openworldassistive}, this constitutes a 4.5\% improvement in detection success rate and a 4.2\% improvement in grasp success rate. 
Table \ref{tab:exp1_time} shows the runtime of our system. On average, each grasp took 51.7s to perform, with almost half the time required to execute the reaching motion.

\subsection{Evaluating Grasping In-Situ}
\label{sec:exp2}

\subsubsection{Experimental Protocol} 

\begin{figure}[t]
    \centering
    \subfloat[]
    {
        \includegraphics[height=1.7in]{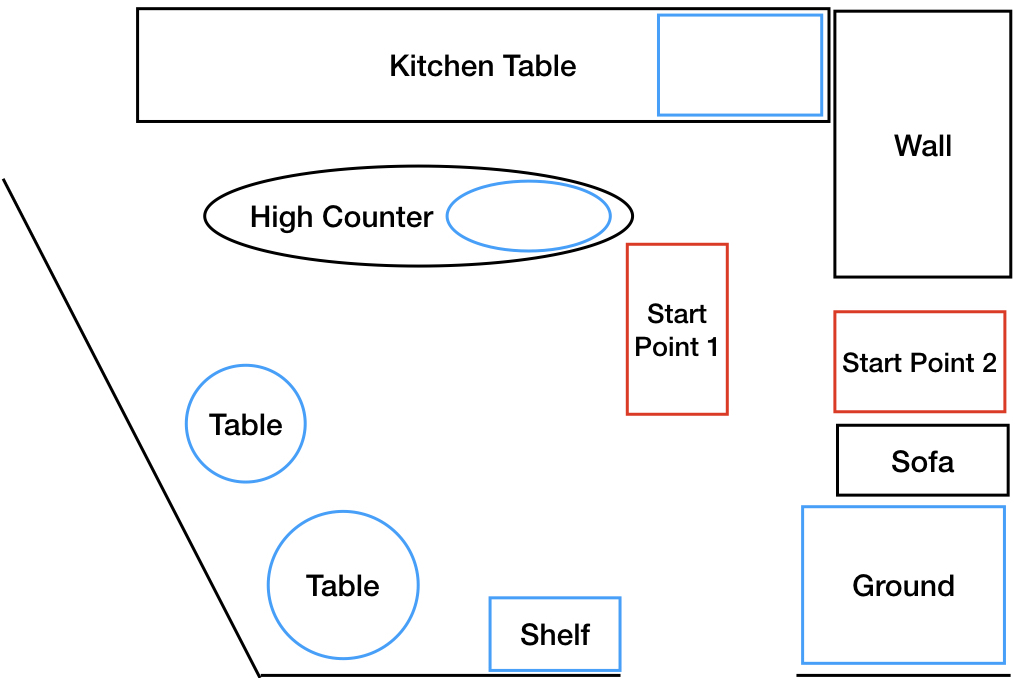}
        \label{setup}
    }
    \subfloat[]
    {
        \includegraphics[height=1.7in]{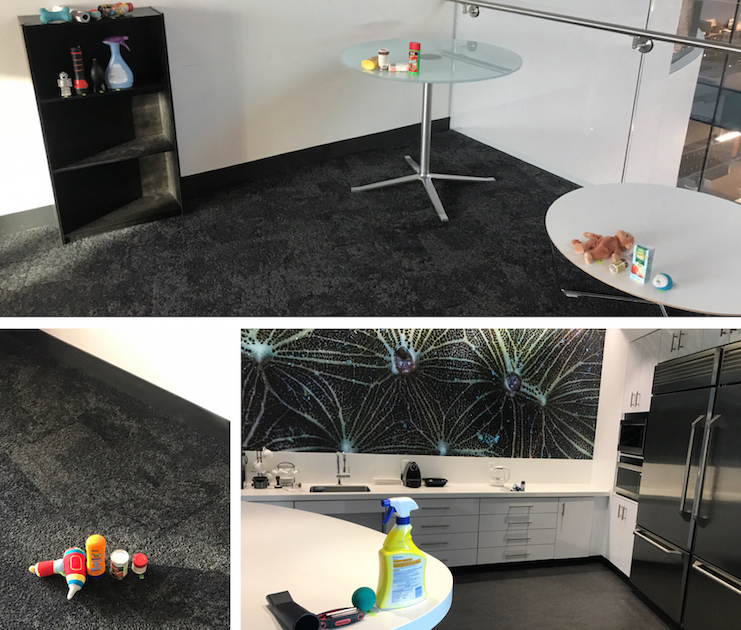}
        \label{example_setup}
    }

    \caption{(a) The kitchen layout for in-situ experiments. (b) One experimental setup in the kitchen environment.}
\end{figure}

To test our system in a real-world scenario, we used the open kitchen area shown in Fig.~\ref{setup}. We performed five trials. In each trial, ten objects were randomly selected from the object set and placed in randomly chosen positions. Three objects were placed on a 73 cm tall table, three were placed on a 31 cm tall table, two were placed on the top shelf of a bookshelf (100 cm high), and two were placed on the middle shelf of the bookshelf (69 cm high). Certain objects were not allowed to be placed on the middle shelf as they could only be grasped from the top down and the arm could not fit inside the shelf. Each trial was run with automatic grasp selection in pick-and-drop mode by an expert user. The sequence in which the objects had to be grasped was randomly generated for each trial. For each object in the sequence, the scooter was first driven to \textit{Start Point 1} (see Fig.~\ref{setup}) and then up to the desired object. We allowed the user to trigger the system several times (until the target object was retrieved). To compare to the work of Gualtieri \etal~\cite{openworldassistive}, we use the same evaluation metrics for this experiment (additional experimental records were provided by the authors of~\cite{openworldassistive}). 

\begin{figure}
\begin{floatrow}
\capbtabbox{%
\scriptsize{
\setlength\tabcolsep{1pt}%
    \begin{tabular}{|c|c|c|}
    \hline
     & {Work of~\cite{openworldassistive}} & {This Work} \\
    \hline
    {Task SR} & 86.0\% (43/50) & 96.0\%(48/50) \\
    \hline
    \multicolumn{3}{c}{} \\
    \hline
    {Failure Type} & {Work of~\cite{openworldassistive}} & {This Work} \\
    \hline
    {Detect Laser} & 7.6\% (5/66) & 10.3\% (7/68) \\
    \hline
    {Lost Track} & 15.1\% (14/93) &  n.a.\\
    \hline
    {Plan Grasp} & 16.5\% (13/79) & 10.3\% (6/58)
    \\
    \hline
    {Wrong Object} & 3.0\% (2/66) & 4.4\% (3/68) \\
    \hline
    {Execute Grasp} & 27.9\% (17/61) & 4.0\% (2/50) \\
    \hline
    \end{tabular}
    }
}{%
    \caption{Results for Grasping In-Situ}
    \label{tab:exp2_result}
}
\capbtabbox{%
\setlength\tabcolsep{1pt}%
    \scriptsize{
    \begin{tabular}{|c|c|c|c|c|}
    \hline
    \multirow{2}{*}{} & \multicolumn{2}{c|}{Work of~\cite{openworldassistive}} & \multicolumn{2}{c|}{{This Work}} \\
    \cline{2-5}
     & {Avg.} & {Std.} & {Avg.} & {Std.} \\
    \hline
    {Acquire Point Cloud} & - & - & 8.6s & 2.7s\\
    \hline
    {Detect Laser Pointer} & - & - & 6.0s & 3.6s \\
    \hline
    {Detect Grasp Pose} & - & - & 4.7s & 1.7s \\
    \hline
    {Filter Grasp} & - & - & 1.6s & 0.9s \\
    \hline
    {Plan Motion} & - & - & 2.9s & 2.9s\\
    \hline
    {Execute} & - & - & 24.3s & 1.1s \\
    \hline
    {Attempt Total} & - & - & 50.7s & 5.7s \\
    \hline
    \hline
    {Driving} & - & - & 11.4s & 4.7s \\
    \hline
    {System Task Total} & - & - & 57.7s & 15.3s \\
    \hline
    {Task Total} & 128s & 99s & 69.0s & 18.4s\\
    \hline
    \end{tabular}
    }
}{%
    \caption{Timing for Grasping In-Situ}
    \label{tab:exp2_time}
}
\end{floatrow}
\end{figure}

\subsubsection{Results}
Out of the 50 tasks (5 trials each with 10 objects), 48 succeeded, giving us a task success rate of 96\%. The two failures were due to objects falling over on the middle shelf during the experiment, making it impossible for the system to grasp them. As is shown in Table~\ref{tab:exp2_result}, our system has a 10\% higher task success rate and a 24\% lower grasp execution failure rate compared to~\cite{openworldassistive}. In 6 of the 58 attempts the system failed to generate a feasible grasp for execution (due to a failure in grasp detection, IK solving, or motion planning), showing a 6.2\% improvement compared to the prior work. While the prior work had a track loss of 15.1\%, our system never has track loss, thanks to our sensing strategy that does not require metric SLAM. However, our system shows a minor increment (2.7\%) in laser detection failure. With respect to wrong object failures (i.e., the robot grasps an object that was not targeted by the user), our system has similar performance as the baseline. Table~\ref{tab:exp2_time} shows the average runtime for this experiment. The average time to complete a successful attempt was 50.7s. When including failed attempts and driving, the time increases to 69s. This is a significant improvement over the 128s average time reported for the baseline~\cite{openworldassistive}.


\subsection{Evaluating Pick-and-Place In-Situ}
\label{sec:exp3}
\subsubsection{Experimental Protocol}
We tested the pick-and-place functionality in a more complex version of the kitchen environment from the previous experiment. Along with the shelf and two tables, there was also a 108 cm tall high table, a 87 cm tall kitchen counter, and a ground area. On each of these seven surfaces we put four numbered tags 7 cm away from each other. We performed three trials. In each trial, 28 objects were randomly sampled from the object set, labeled from 1 to 28, and placed on top of the corresponding tag. The orientation of the objects was determined by a fair coin flip (1=upright, 0=not upright). Fig.~\ref{example_setup} shows a setup of one trial. Each object was randomly assigned a target surface and the order the objects were to be pick-and-placed in was determined by randomly generating a permutation of 1 to 28. These trials were run with manual grasp selection in pick-and-place mode by an expert user. At the start of each task, the scooter was driven to \textit{Start Point 2} (seen in Fig.~\ref{setup}) and then up to the next object to be grasped. After the object was grasped, the scooter was driven to the target surface and the object was placed on top of it. During the experiment, the user was allowed to perform pick-and-place actions to separate the target object from nearby objects.
We evaluated the same metrics as in Section~\ref{sec:exp2}. 
Fig. \ref{fig:exp_open} shows our system grasping an object in this experiment.

\begin{figure}[t]
\begin{floatrow}
\capbtabbox{%
\scriptsize{
\setlength\tabcolsep{1pt}%
    \begin{tabular}{|c|c|c|}
    \hline
    {Task SR} & \multicolumn{2}{c|}{96.4\%(81/84)} \\
    \hline
    \multicolumn{3}{c}{} \\
    \hline
    Failure Type & {Grasping} & {Placing} \\
    \hline
    {Detect Laser} & 2.0\%(3/150) & 1.0\%(1/104) \\
    \hline
    {Detect Grasp} & 8.0\%(12/150) & - \\
    \hline
    {IK/Planning} & 20.0\%(27/135) & 7.8\%(8/103) \\
    \hline
    {Wrong Object} & 0\%(0/135) & - \\
    \hline
    {Execute} & 12.0\%(13/108) & 3.2\%(3/95) \\
    \hline
    \end{tabular}
    }
}{%
    \caption{Results for Pick-and-Place In-situ}
    \label{tab:exp3_result}
}
\capbtabbox{%
\setlength\tabcolsep{1pt}%
    \scriptsize{
    \begin{tabular}{|c|c|c|c|c|}
    \hline
    & \multicolumn{2}{c|}{Avg.} & \multicolumn{2}{c|}{Std.} \\
    \hline
    {System Task Total} & \multicolumn{2}{c|}{129.9s} & \multicolumn{2}{c|}{75.6s}\\
    \hline
    \multicolumn{5}{}{} \\
    \hline
    \multirow{2}{*}{} & \multicolumn{2}{c|}{{Grasping}} & \multicolumn{2}{c|}{{Placing}} \\
    \cline{2-5}
     & {Avg.} & {Std.} & {Avg.} & {Std.} \\
    \hline
    {Acquire Cloud} & 8.3s & 2.5s & 9.5s & 4.4s \\
    \hline
    {Detect Laser Pointer} & 6.0s & 2.2s & 4.7s & 3.0s \\
    \hline
    {Detect Grasp Pose} & 3.7s & 1.6s & - & - \\
    \hline
    {Grasp Filter} & 4.9s & 2.6s & - & - \\
    \hline
    {Grasp Selection} & 2.5s & 2.6s & - & - \\
    \hline
    {Motion Planning} & 7.5s & 6.4s & 1.1s & 5.0s \\
    \hline
    {Execution} & 17.1s & 3.9s & 15.6s & 1.5s \\
    \hline
    {Attempt Total} & 53.3s & 9.5s & 35.1s & 10.2s \\
    \hline
    \end{tabular}
    }
}{%
    \caption{Timing for Pick-and-Place In-Situ}
    \label{tab:exp3_time}
}
\end{floatrow}
\end{figure}

\subsubsection{Results}
Out of the 84 tasks (3 trials with 28 tasks each trial), 81 succeeded giving us a 96.4\% task success rate. In 13 of the successful tasks, it was necessary to move a nearby object out of the way (we did this by performing a separate pick/place with the scooter) in order to reach the target object. 
As is shown in Table \ref{tab:exp3_result}, the laser detection failure drops compared to the previous experiment because this experiment was conducted at night and there was less interference caused by sunlight. The grasp detection failure rate remains similar to the previous experiment, but the system suffered from more IK/planning failures and execution failures. The primary reason is the increment in complexity of the environment. The high counter top is especially challenging because it nearly reaches the arm's workspace limit. 
No wrong object failure appeared because of the use of manual grasp selection. This enabled the user to select on-target grasps.
The average time needed for each pick-and-place task was 130s excluding time for driving. Note that this is higher than the sum of the total grasping time and the total placing time in Table~\ref{tab:exp3_time} because it was not guaranteed that each task could be finished by only one pick-and-place attempt.

\subsection{Comparing Automatic and Manual Grasp Selection}
\subsubsection{Experimental Protocol}

Finally, we compared the automatic grasp selection and manual grasp selection. We performed 50 trials with the same table setup as in Section~\ref{sec:exp1}. In each trial, two objects were randomly selected and placed next to each other on the table. One of the two objects was randomly chosen as the target object, and an expert user operated the system to perform one pick-and-drop on the target object with automatic grasp selection mode and manual grasp selection mode, respectively. Note that for each mode, there was only one attempt allowed to pick up the target object. We evaluated the wrong object and the grasp execution failure rates of the two grasp selection modes.

\begin{table}[t]
\begin{center}
\scriptsize{
\begin{tabular}
{|c|c|c|}
\hline
& {Automatic Selection} & {Manual Selection} \\
\hline
{Wrong Object Failure} & 26.0\% (13/50) & 8.0\% (4/50) \\
\hline
{Grasp Execution Failure} & 6.0\% (3/50) & 14.0\% (7/50)\\
\hline
\end{tabular}
}
\end{center}
\caption{Results of Comparing Automatic and Manual Grasp Selection}
\label{tab:exp4}
\end{table}

\subsubsection{Results}
As shown in Table~\ref{tab:exp4}, manual grasp selection has a 20\% lower wrong object failure rate because it allows the user to choose on-target grasps when applicable. This eliminates the possibility that the system selects a grasp to be executed on an object that is not the target but had higher objective function value, as is the case with automatic grasp selection. However, there were still three occasions for the manual selection mode where our system only provided grasps on the wrong object. Interestingly, automatic grasp selection shows a 6\% lower grasp execution failure rate. The main reason for this is that the robot was forced to execute a challenging grasp on a target object (i.e., a grasp that could easily result in a collision), while automatic grasp selection simply executed easier grasps on the wrong object.

\section{DISCUSSION AND LIMITATIONS}

We present an assistive robotic system that enables pick and place of novel objects in open world environments. Our experimental results demonstrate reliable pick-and-drop and pick-and-place capabilities. In pick-and-drop in-situ experiments, we reached a grasp execution success rate of 96\% and an average total runtime of 57.7s. In pick-and-place in-situ experiments, we reached a grasp execution success rate of 88\%, a place execution success rate of 96.8\%, and an average total runtime of 129.9s.

The two most common failure modes of our system are: (i) not immediately finding a feasible motion plan, and (ii) not detecting the desired object with the laser pointer. The first failure mode could be addressed by using a different planner. The second failure mode could be addressed by developing a better laser pointer detection algorithm. One way to do this might be to use RGBD sensors, such as the Intel Real Sense, to be able to exploit the RGB. That should facilitate the detection of the laser pointer.

The runtime of our system could be reduced in multiple ways. Currently, point cloud acquisition, laser pointer detection, grasp pose detection, and robot arm motions take up most of the time. To avoid infrared interference that causes the point clouds to be very noisy and have a large amount of missing depth data, we currently switch the sensors on and off, which takes a lot of time. This problem could be tackled by using stereo cameras which do not use infrared and therefore are not required to be turned on and off. 
Grasp pose detection could be made faster by using a CPU with more cores, enabling a larger degree of parallel computations. 

The pick-and-place problems that our system addresses are relatively simple. When placing, we only consider the target position of the placement and mainly ignore the object's orientation. In the future, we would like to allow the user to choose a target orientation for the placement and enhance our system with the capability to reorient the object accordingly. Furthermore, we can currently not maintain a fixed object orientation while moving the robot arm after grasping. This would be a problem if the robot had to pick up an object filled with a liquid, e.g., a cup of coffee.



Our user interface provides another avenue for future research. 
Currently the projector is only used for displaying the next target of our system, however there are a number of ways to expand its usage to display more information to the user. For example, it could display the locations of planned grasps or the workspace of the arm. Our objective is to remove the screen from the system, allowing people to give commands to and receive information from the system in the real world.

User studies are another direction for future research. As this system aims to assist people with different abilities, we first plan to develop a variety of adaptations of user interface with different access methods. User studies could then be used to determine which interfaces work the best overall and which interfaces work better for specific demographics of users. This would allow us to provide a robust assistive system which is applicable to a large variety of users.

\section*{ACKNOWLEDGEMENTS}
\begin{sloppypar}
This work has been supported in part by the National Science Foundation (IIS-1426968, IIS-1427081, IIS-1724191, IIS-1724257, IIS-1763469), NASA (NNX16AC48A, NNX13AQ85G), ONR (N000141410047), Amazon through an ARA to Platt, and Google through a FRA to Platt.
\end{sloppypar}

%
%

\end{document}